\documentclass{article} 
\usepackage{nips13submit_e,times}
\usepackage{url}

\usepackage{amsmath}
\usepackage{amssymb}
\DeclareMathOperator{\tr}{tr}
\usepackage{tabularx}

\usepackage{graphicx} 

\usepackage{algorithm}
\usepackage{algorithmic}

\usepackage{subfig} 
\graphicspath{{images/}}

\title{Controlling the Precision-Recall Tradeoff in Differential
           Dependency Network Analysis}

\author{
Diane Oyen\\
University of New Mexico\\
Albuquerque, NM USA \\
\texttt{doyen@cs.unm.edu} \\
\And
Alexandru Niculescu-Mizil \\
NEC Laboratories \\
Princeton, NJ USA \\
\texttt{alex@nec-labs.com} \\
\AND
Rachel Ostroff \\
SomaLogic Inc. \\
Boulder, CO USA \\
\texttt{rostroff@somalogic.com} \\
\And
Alex Stewart \\
SomaLogic Inc. \\
Boulder, CO USA \\
\texttt{astewart@somalogic.com} \\
\And
Vincent P. Clark \\
University of New Mexico \\
Mind Research Network \\
Albuquerque, NM USA \\
\texttt{vclark@unm.edu} \\
}

\nipsfinalcopy 

\begin{document} 

\maketitle

\begin{abstract} 

Graphical models have gained a lot of attention recently as a tool for
learning and representing dependencies among variables in multivariate
data. Often, domain scientists are looking specifically for
differences among the dependency networks of different conditions or
populations (e.g. differences between regulatory networks of different
species, or differences between dependency networks of diseased versus
healthy populations). The standard method for finding these
differences is to learn the dependency networks for each condition
independently and compare them. We show that this approach is prone to
high false discovery rates (low precision) that can render the
analysis useless.  We then show that by imposing a bias towards
learning similar dependency networks for each condition the false
discovery rates can be reduced to acceptable levels, at the cost of
finding a reduced number of differences. Algorithms developed in the
transfer learning literature can be used to vary the strength of the
imposed similarity bias and provide a natural mechanism to smoothly
adjust this differential precision-recall tradeoff to cater to the
requirements of the analysis conducted.  We present real case studies
(oncological and neurological) where domain experts use the proposed
technique to extract useful differential networks that shed light on
the biological processes involved in cancer and brain function.

\end{abstract} 

\section{Introduction}

Network structure learning algorithms, such as Gaussian graphical
models, enable scientists to visualize dependency structure in
multivariate data. Recently, attention has been brought to the problem
of identifying differences between the dependency networks of various
conditions or populations. For example, in a neuroimaging study, we
want to understand how regions of the brain share information before
and after a person acquires a particular skill. The goal is to
identify the regions of the brain that are most influential after a
skill has been learned so that direct current stimulation can be
applied to those regions to accelerate a person's learning
process. In another example we analyze how the dependency structure of
plasma proteins changes between healthy patients and patients that
have cancer, with the goal of understanding the cancer biology and
identifying better diagnostics.

Tackling these problems, we found that traditional methods for
differential dependency network analysis, based on learning the
dependency network for each condition independently and then comparing
them, tend to produce a large number of spurious differences. This
hampers the analysis and prevents drawing any reliable conclusions,
significantly limiting the usefulness of the differential analysis.
We also found that there is a need for an intuitive mechanism to
control the quality of the learned differences, and trade off having a
small number of spurious differences (high differential precision)
with identifying a large number of differences (high differential
recall).



In this paper, we propose a novel use of \emph{transfer learning} to
control the precision-recall tradeoff in \emph{differential} network
analysis, and show that this approach dramatically improves the
quality of the learned differences. The key idea is to learn the
dependency networks for the different conditions jointly, imposing a
bias that the learned networks be similar. The more heavily this bias
is enforced, the fewer differences will be learned between
networks. Our thesis is that true differences that are well supported
in the data tend to require a higher bias to be eliminated, while
spurious differences are eliminated with a lower bias. Thus, by
adjusting the strength of the similarity bias, spurious differences
can be filtered out decreasing the number of false discoveries and
increasing the reliability of the analysis. Using this technique in
two oncology studies we identify differential dependencies that give
insight into cancer biology. In a neuroimaging study we find known
visual processing pathways and discover interesting insights into
regions that relate to visual object recognition.

{\bf Related Work.}  The most common method for performing
differential network analysis is to learn the networks independently
for each condition and compare them. As a post-hoc analysis, a
bootstrap procedure or permutation test can be applied to eliminate
some of the false differences (e.g. \cite{zhang2009differential}). We
show that the transfer learning based approach performs significantly
better and is far less computationally expensive than the bootstrap
procedure. A related, but different problem is to learn (Bayesian)
networks that discriminate between conditions
(e.g. \cite{grossman_learning_2004}). Discriminative methods introduce
an arc between two variables when their interaction gives useful
discriminatory information. This does not mean, and usually is not the
case, that there is a statistical dependency between the two variables
in either condition.

Transfer learning algorithms for graphical models have been
extensively studied, and have been shown to produce networks that are
more accurate than networks learned independently
\cite{danaher_joint_2011,Chiquet:2011:IMG:2036043.2036078,guo2011joint}. However,
we are not aware of any existing research that investigates using
transfer learning to obtain high quality \emph{differences between
  networks} or to provide a mechanism to control the precision-recall
tradeoff in \emph{differential} network analysis.  Danaher et
al.~\cite{danaher_joint_2011}, mention low recall of differences
learned on synthetic data, but do not explore further. A recent paper
\cite{NIPS2012_0291} explores techniques for biasing learning such
that the dependency networks differ in a limited number of
variables. They show that if the differences match their assumption,
the individual networks can be recovered more accurately.  We
emphasize that our interest lies squarely in improving the quality of
the \emph{differential} dependency network analysis and providing an
intuitive mechanism for trading off the precision and recall of the
learned differences. Improving the quality of individual dependency
networks, or devising new algorithms for dependency network or
transfer learning, while interesting, are orthogonal to the scope of
this paper.

\section{Learning Independent Networks then Comparing Will Not Work}

\begin{figure}[tb]
\centering
\subfloat[]{\includegraphics[width=0.3\columnwidth]{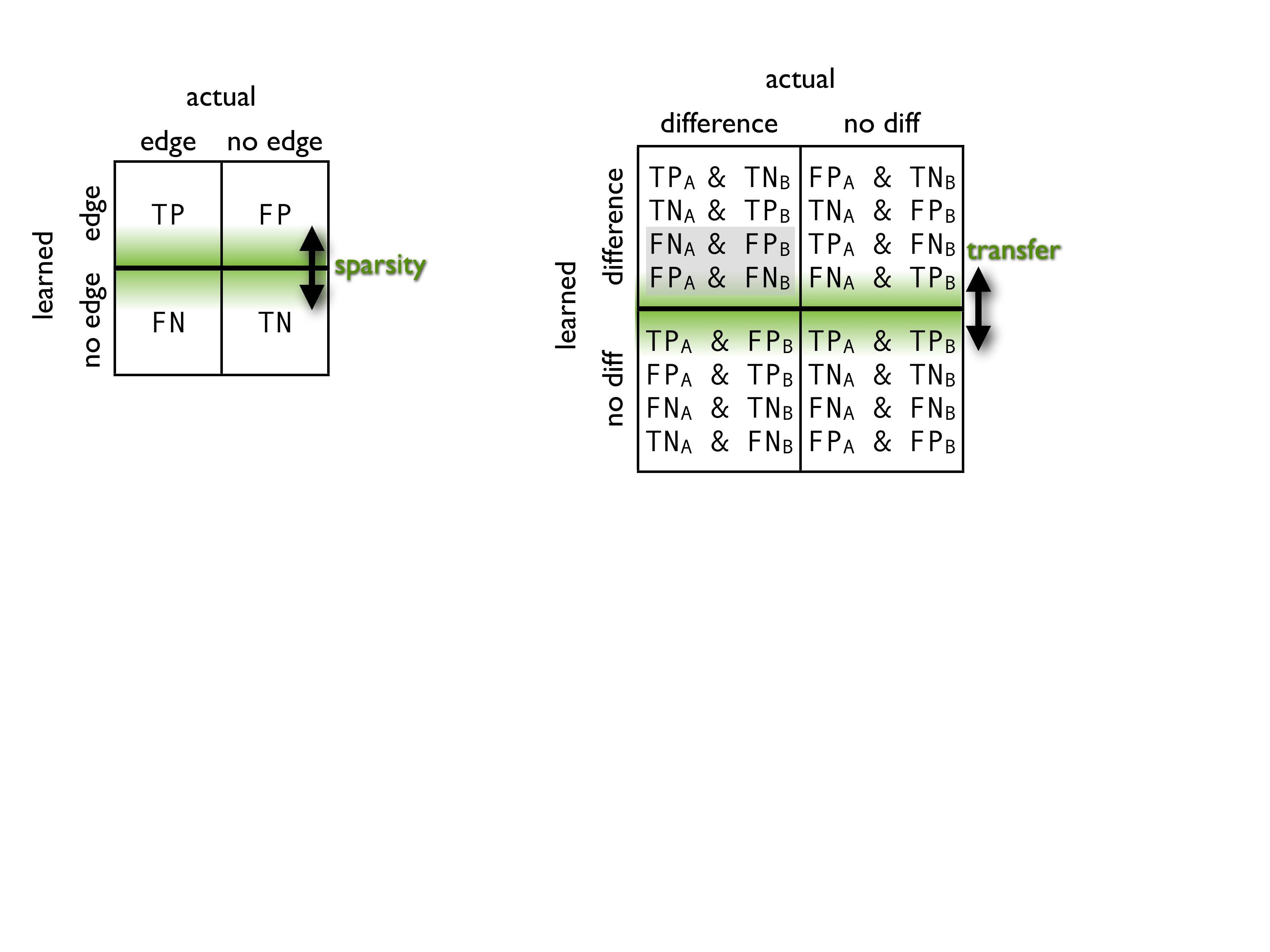}\label{fig:confusionMatrixEdge}}
\hfil
\subfloat[]{\includegraphics[width=0.4\columnwidth]{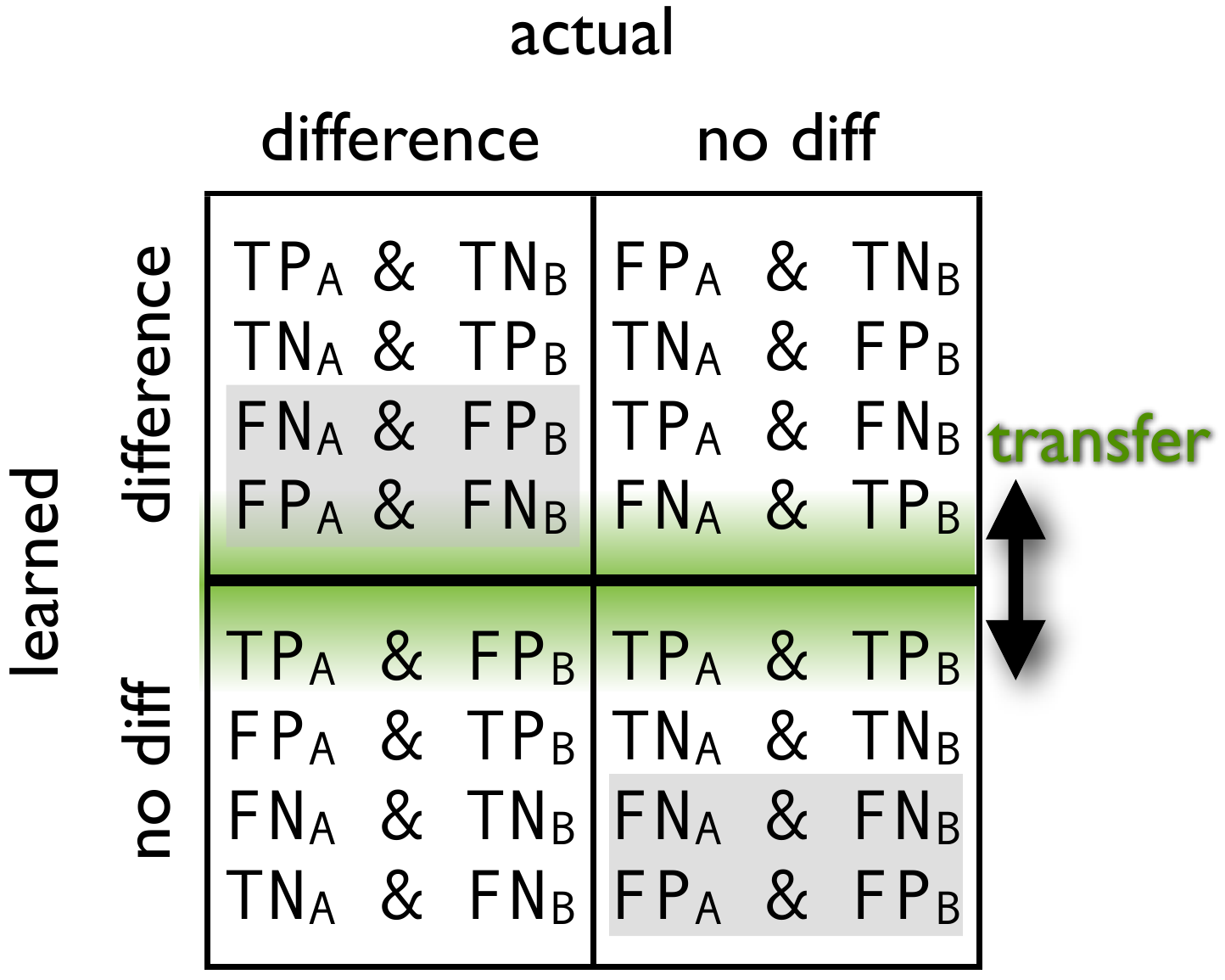}\label{fig:confusionMatrixDiff}}
\\
\subfloat[]{\includegraphics[width=0.48\columnwidth]{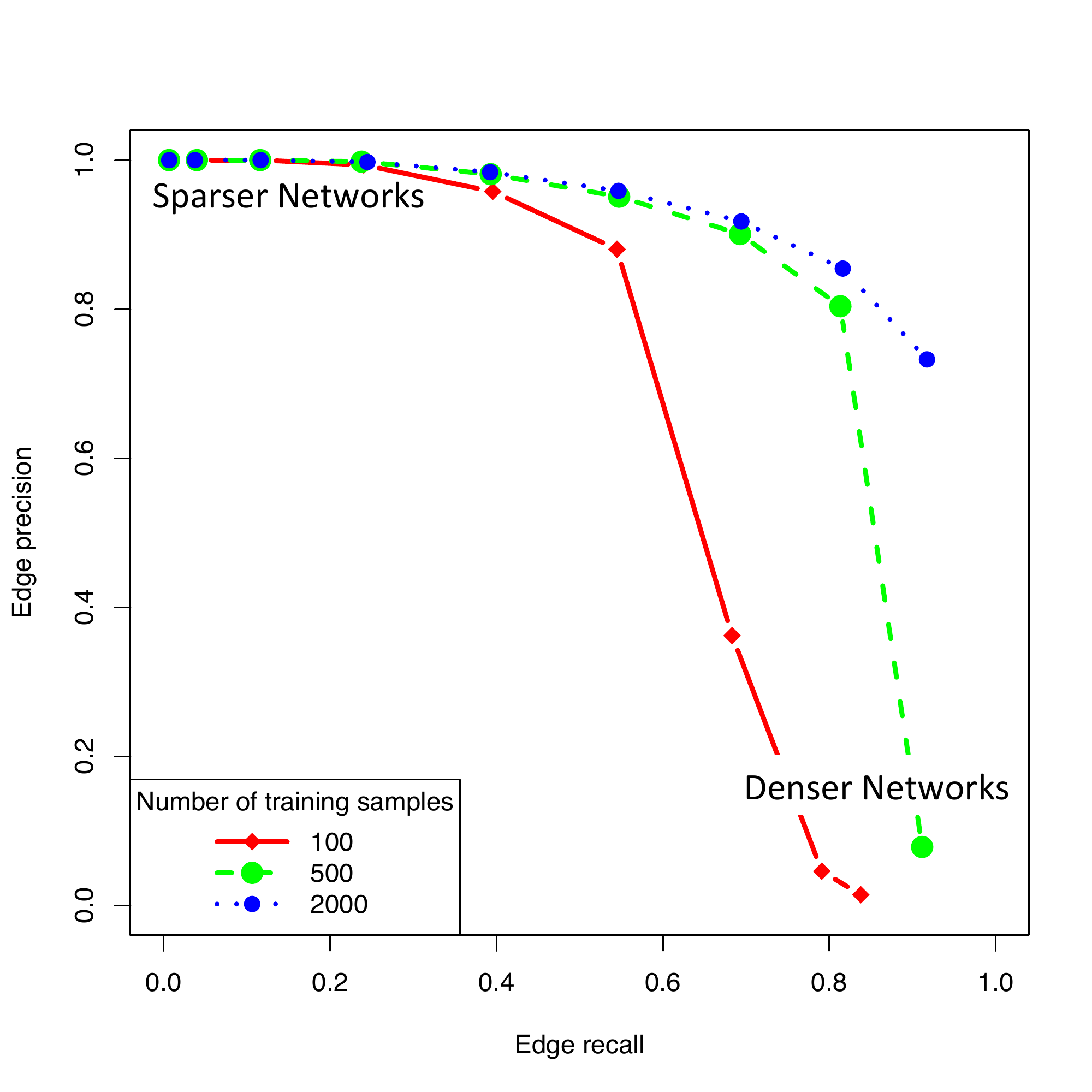}\label{fig:edge_pr}} \hfil
\subfloat[]{\includegraphics[width=0.48\columnwidth]{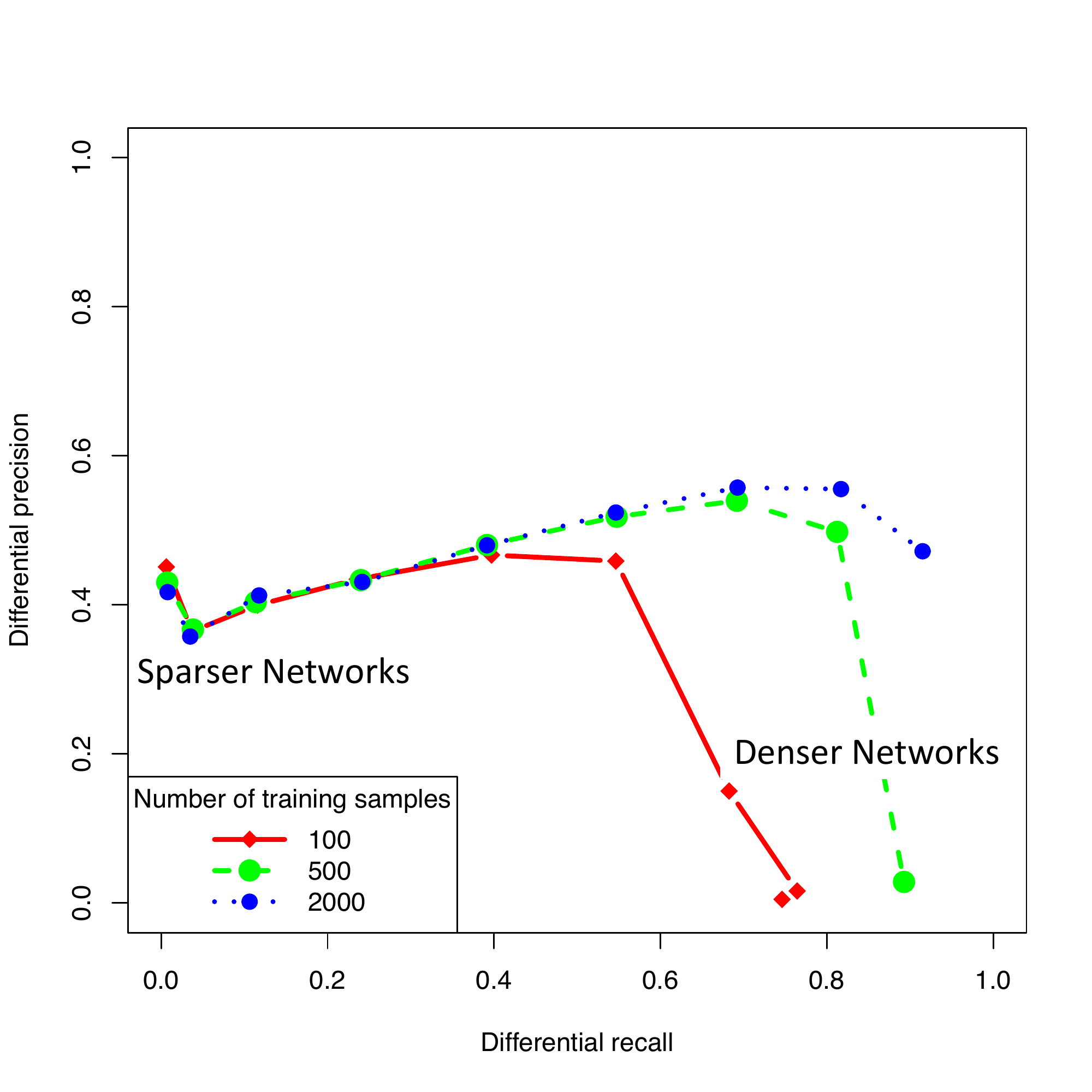}\label{fig:diff_pr}}
\caption{(a) Confusion matrix for learning edges of a single network, (b) Confusion matrix for learning differences between networks A and B, (c) Edge precision-recall graph, (d) Differential precision-recall graph.}


\label{fig:confusionMatrix}
\end{figure}

In real applications data is often limited and noisy, and modeling
assumptions usually do not hold. So we must assume that there will be
errors when learning dependency networks from data. The different
types of errors can be visualized in a confusion matrix, as in
Figure~\ref{fig:confusionMatrixEdge}. Ideally, all edges would be
identified as true positives (TP) or true negatives (TN), but this is
usually not possible so there will also be some false positives (FP)
and false negatives (FN). Using sparse network learning algorithms, such
as graphical lasso \cite{meinshausen_high-dimensional_2006}, one can
trade off between the two types of errors by adjusting the degree of
sparsity of the learned network. This moves the boundary between the
\emph{learned} edges and non-edges shown as the horizontal line
highlighted in green in Figure~\ref{fig:confusionMatrixEdge}. Assuming
the algorithm is able to identify edges with better than random
probability, the \emph{precision} ($\text{TP} / (\text{TP} +
\text{FP})$) will increase with sparsity; meanwhile, the \emph{recall}
($\text{TP} / (\text{TP} + \text{FN})$) will
decrease. Figure~\ref{fig:edge_pr} shows that this is indeed the
case. The figure plots the edge recall vs. the edge precision for
networks learned from training sets of various sizes. The true network
has 1000 nodes and 1000 edges. Each line is generated by changing the
sparsity to obtain different edge precision-recall tradeoffs. For
denser networks, recall is high, but precision is lower. As networks
get sparser, the precision increases but the recall decreases.  Thus
the degree of sparsity controls the edge precision-recall tradeoff.

When conducting a differential network analysis (i.e. identifying
differences between networks), one would like to control in a similar
manner the tradeoff between the \emph{differential precision} (the
percentage of the inferred differences that are actually true) and the
\emph{differential recall} (the percentage of true differences that
are recovered). Unfortunately, the traditional approach of learning
the networks independently and comparing them provides no mechanism
for controlling the differential precision-recall tradeoff. Also,
adjusting the sparsity of the learned networks will not help with
obtaining the high differential precision required in many
applications (e.g. in biological applications, a differential
precision above 80\% (FDR of 20\%) is
desirable). Figure~\ref{fig:diff_pr} plots the {\em differential}
precision versus the {\em differential} recall for pairs of networks
learned from training sets of various sizes. The true networks have
1000 nodes and 1000 edges and about 80\% of edges in common. Each line
is obtained by varying the sparsity of the learned networks. While
adjusting the sparsity has some influence on the differential
precision and recall, the differential precision never gets above
60\%. The reason for this is that the sparsity controls a tradeoff
between two types of mistakes: inferring an edge where no edge exists,
and missing a true edge (FP and FN in
figure~\ref{fig:confusionMatrixEdge}). Both types of mistakes will
lower the differential precision if the other network does not make
the same mistake (see Figure~\ref{fig:confusionMatrixDiff}), so
trading off between them will not improve the differential precision.

It is important to note that getting more data
will not solve this problem. Increasing the training set size four
fold from 500 to 2000 instances per task barely improves the
differential precision. Also note that the learning algorithm does a
very good job at recovering the individual networks
(Figure~\ref{fig:edge_pr}). Thus, short of learning almost perfect networks
which is usually impossible in practice, simply improving the
performance of the individual network learning algorithms will not be enough to
obtain the high differential precision required in many practical
applications.

\section{Obtaining a Differential Precision-Recall Tradeoff}

The differential confusion matrix in
Figure~\ref{fig:confusionMatrixDiff} provides a clue about how to
obtain the desired differential precision-recall tradeoff: the
horizontal line can be controlled by imposing an inductive bias
towards learning similar networks. A stronger bias for similar
networks leads to fewer \emph{learned} differences, while a weaker
bias leads to more \emph{learned} differences. Assuming that true
differences can better overcome this bias, the differential precision
will increase with a stronger bias, while the differential recall will
probably decrease. Thus the strength of the bias for similar networks
is controlling the differential precision-recall tradeoff much in the
same way the sparsity bias controls the edge precision-recall
tradeoff.

To impose an inductive bias towards learning similar networks, we
borrow techniques developed in the transfer learning literature. In
transfer learning or multi-task learning, inductive bias towards
similar networks is used to obtain more accurate dependency structures
when the true networks are similar
\cite{danaher_joint_2011,Chiquet:2011:IMG:2036043.2036078,guo2011joint}.
The same algorithms can be employed to control the differential
precision-recall tradeoff. We emphasize that, even though the learning
algorithm is the same, the goal is different. In transfer learning the
goal is to improve the accuracy of the individual networks while in
this paper the goal is to improve the differential precision and
control the differential precision-recall tradeoff. As we shall see,
this different goal makes the technique more widely applicable and
easier to use. For instance we do not need to assume that the true
networks are similar. 

In this paper, we use the joint graphical lasso algorithm from
\cite{danaher_joint_2011}, which we very briefly describe
below. However, other transfer learning algorithms can be used as
well. Assuming that the differential analysis is performed over $K$
conditions or populations, the algorithm infers a precision matrix
$\widehat{\Theta}_k$ for each condition by solving the following joint
optimization problem:
\begin{equation*}
\arg\max_{\Theta_k \succ 0, \forall k} \sum_{k=1}^K \left[ \log\det\Theta_k - \tr(\widehat{\Sigma}_k \Theta_k) \right] - \\
\lambda_1 \sum_{i\neq j} \left[ (1 - \lambda_2)  \sum_{k=1}^K |\theta_{k,ij}| + \lambda_2 \left(\sum_{k=1}^K \theta_{k,ij}^2\right) ^{1/2} \right]
\end{equation*}
where $\widehat{\Sigma}_k$ is a generalized correlation matrix
estimated from the data. If Gaussian covariance is used as a measure
of correlation, then a multi-variate Gaussian distribution is fitted
to the data of each condition. In this paper we measure correlation using
Kendall's Tau which expands the model class to transelliptical
graphical models and leads to increased robustness to outliers and
non-Gaussianity without a significant loss in performance
\cite{NIPS2012_0380}. After $\widehat{\Theta}_k$ are learned, a
dependency network for each domain is obtained by connecting all the
variables that have a non-zero entry in $\widehat{\Theta}_k$, and
differences between domains are obtained by comparing these networks.

The parameter $\lambda_1$ controls the sparsity bias for the learned
networks, while $\lambda_2$ ($0\leq \lambda_2 \leq 1$) is the
\emph{similarity bias} parameter and controls the strength of the bias
towards learning similar networks. When $\lambda_2 = 0$, there is no
bias towards similar networks and is equivalent to the
traditional method of learning a network for each condition
independently. As $\lambda_2$ approaches $1$, the
bias towards learning similar networks gets stronger, and only
differences that are highly supported in the data survive. At
$\lambda_2 = 1$ the learned structures will be identical and no
differences will be recovered.

\section{Experiments with Synthetic Data}

We first test the approach using synthetic networks and data. To
create a synthetic data set, we generate a network with 1000 Gaussian
variables and 1000 undirected edges. Then, the endpoint of each edge
is re-wired with some probability to another node, creating a
different network with edges in common with the first one. The goal is
to correctly identify the differences between the two networks. For
each network $k$ we generate a precision matrix $\Theta_k$ by
independently sampling each entry that corresponds to and edge from a
normal distribution, then re-scaling to ensure that $\Theta_k$ is
positive definite. Training data is then drawn from
$\mathcal{N}(0,\Theta_k^{-1})$ for each condition $k$. Results are
averaged across 5 trials.

\begin{figure}[tb]
\centering
\subfloat[Various sample sizes]{\includegraphics[width=0.48\textwidth]{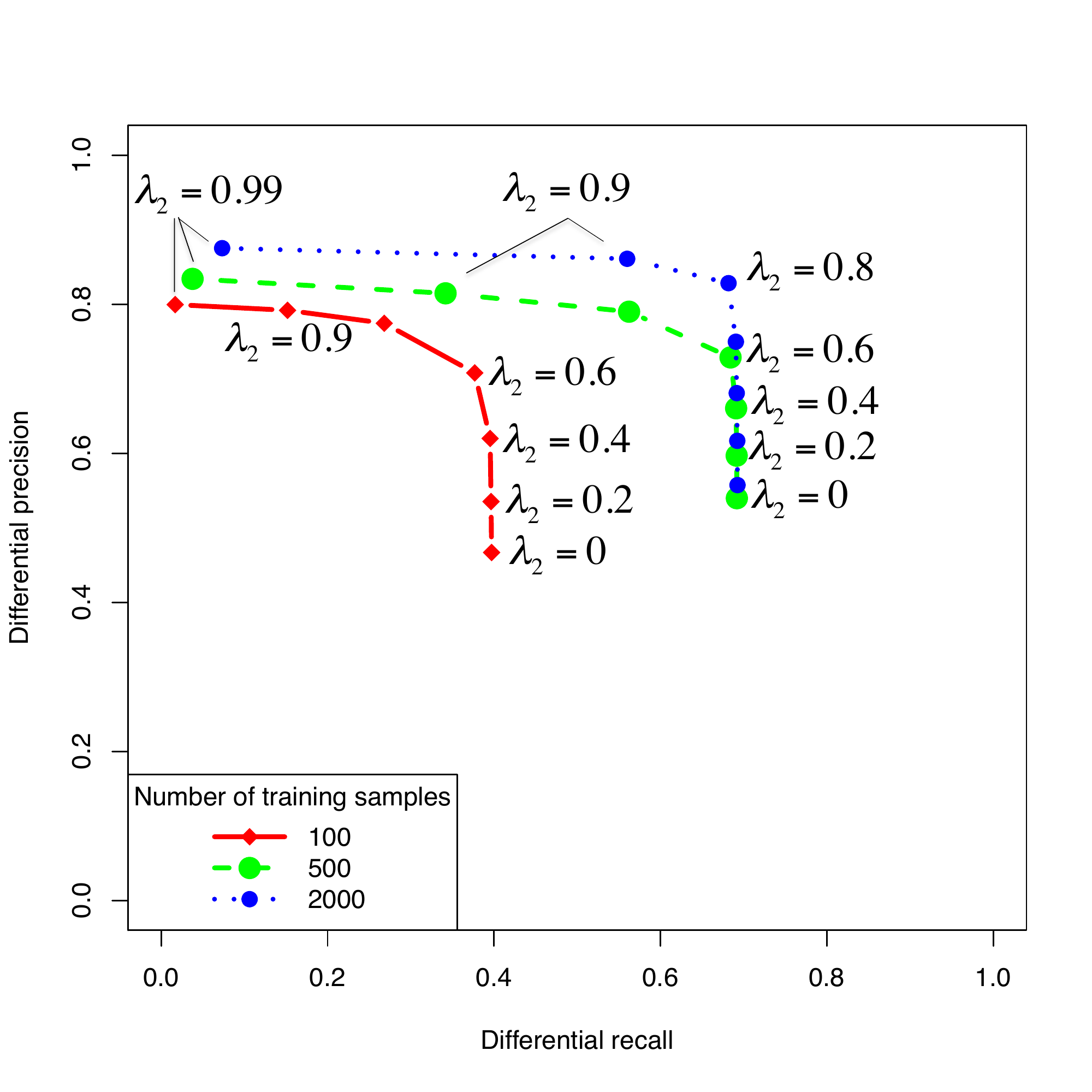}\label{fig:train_size}} \hfil
\subfloat[Various sparsity levels]{\includegraphics[width=0.48\textwidth]{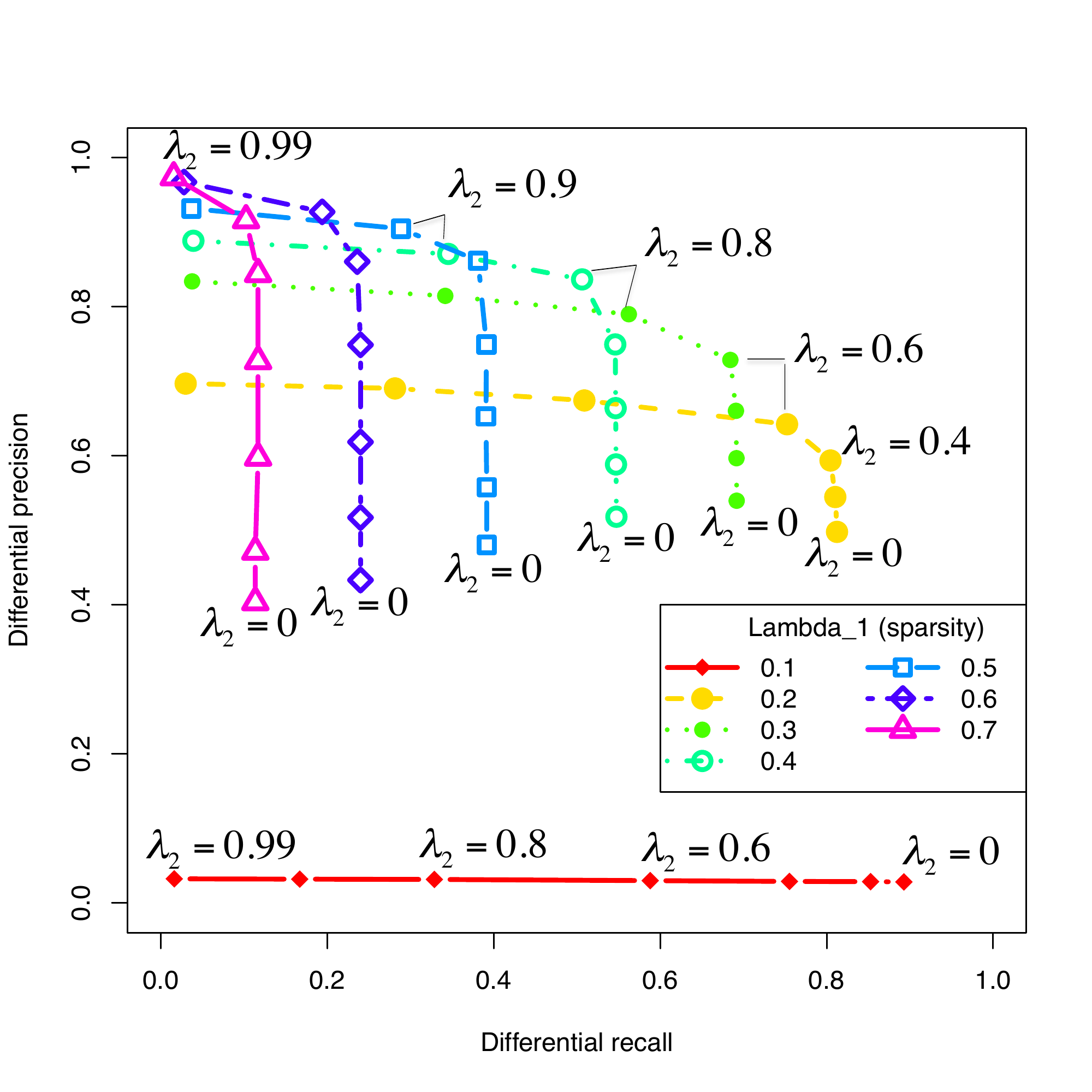}\label{fig:sparsity}}
\\
\subfloat[Various rewiring probabilities]{\includegraphics[width=0.48\textwidth]{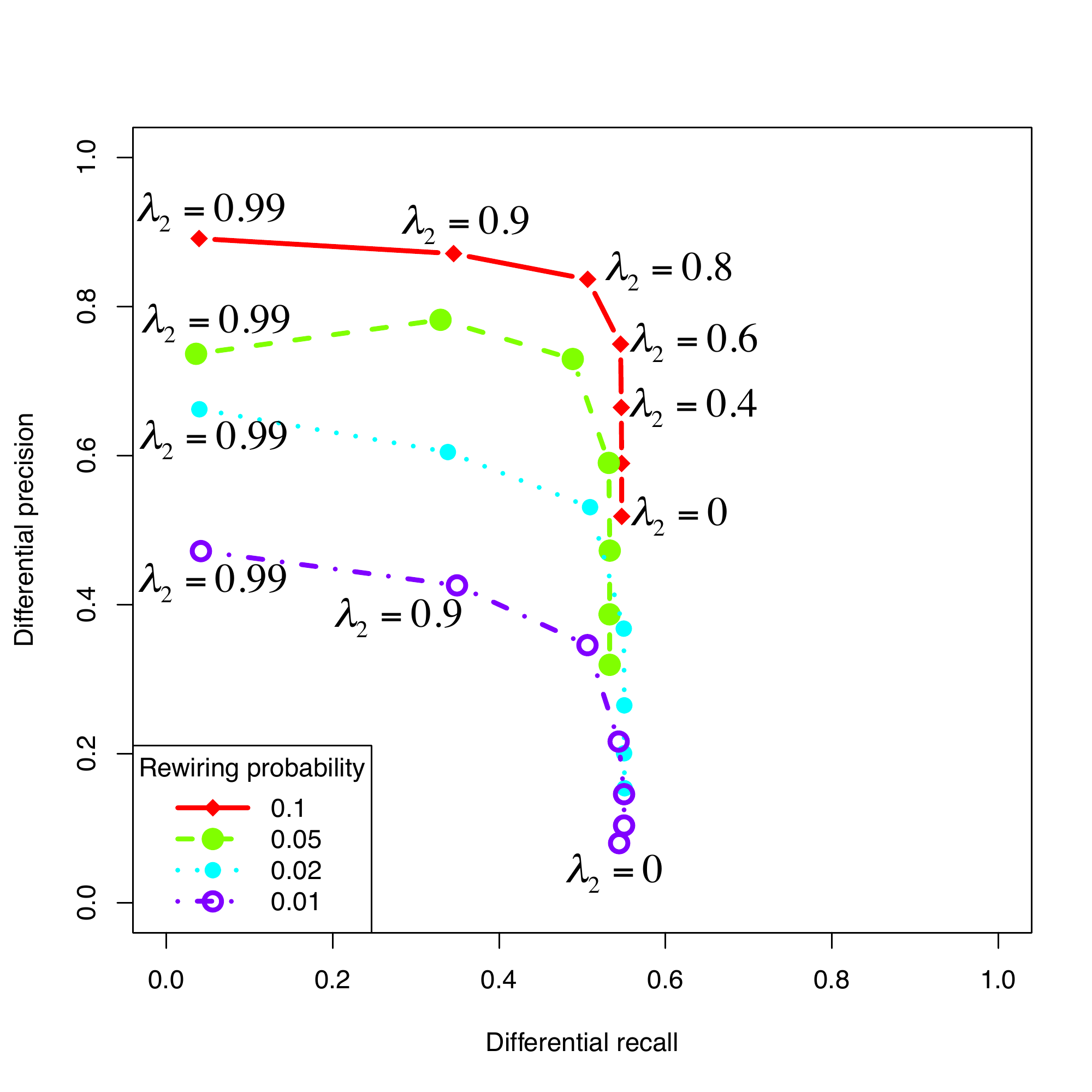}\label{fig:pmove}}
\hfil
\subfloat[Transfer vs. standard bootstrap]{\includegraphics[width=0.48\textwidth]{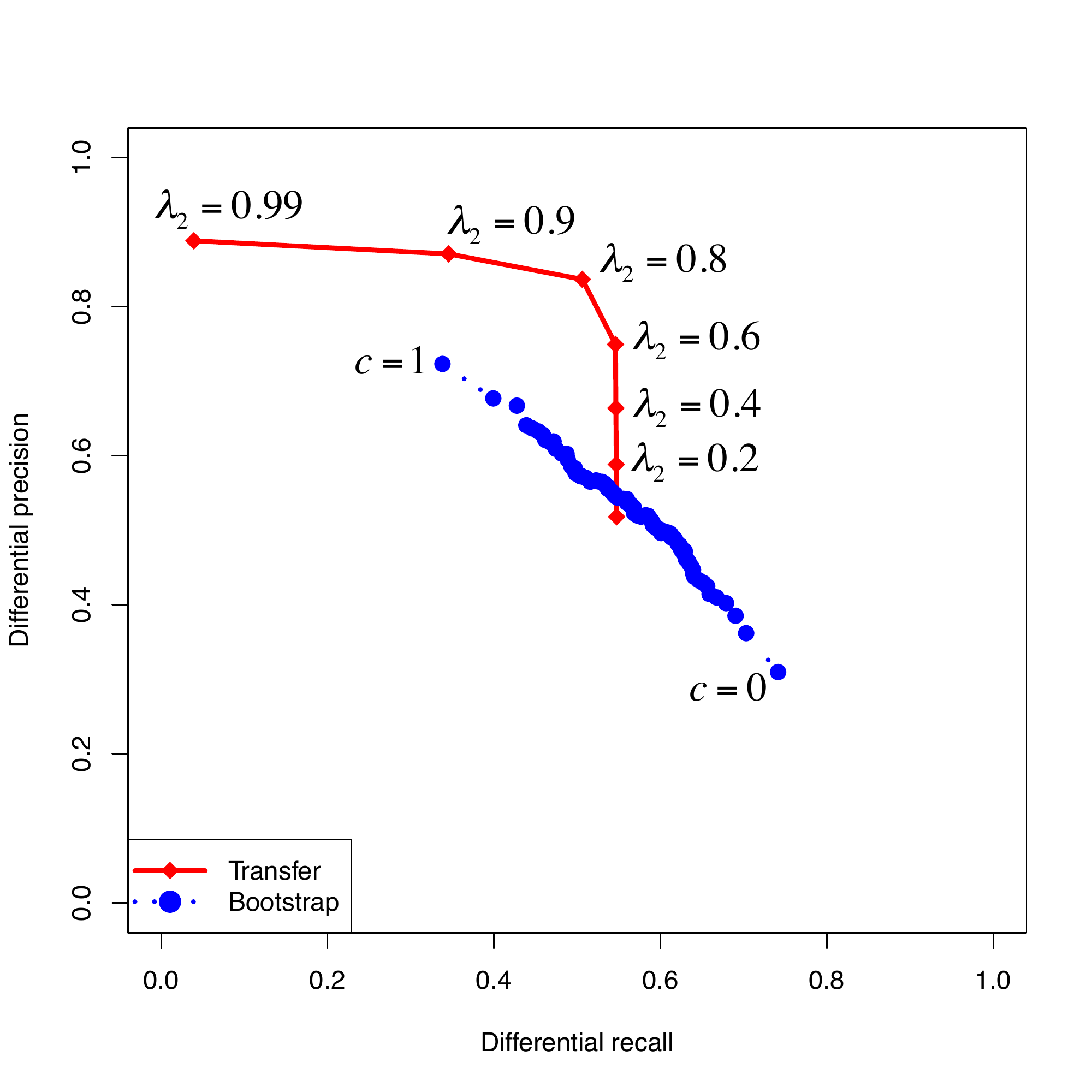}\label{fig:bootstrap}}
\caption{Differential precision-recall curves on synthetic data.}
\label{fig:synthetic_precision}
\end{figure}

The results of the experiments are depicted in
Figure~\ref{fig:synthetic_precision} in terms of differential
precision-recall curves. In these plots, the differential
precision-recall curves are obtained by varying the similarity bias
parameter $\lambda_2$ between 0 and 1 to obtain different tradeoffs
between differential precision and recall. For $\lambda_2=0$ we
recover the performance of the traditional approach of learning the
networks independently and comparing them. This is always the
rightmost point of each differential precision-recall curve, with the
highest recall but the lowest precision.

Figure~\ref{fig:train_size} shows the differential precision-recall
curves for different training set sizes. For each training set size,
the sparsity parameter $\lambda_1$ is set to the value that yields the
highest differential precision in Figure~\ref{fig:diff_pr} (i.e. the best
differential precision obtained by learning the networks
independently and comparing them). For all data sizes, increasing the
similarity bias (increasing $\lambda_2$) improves the differential
precision, showing that the proposed technique does indeed enable a
more reliable differential analysis. Even with as little data as 100
instances per condition, we are able to obtain a differential
precision above 0.8 which would be considered acceptable in many
applications. The price to pay is a reduction in the number of
differences recovered (reduction in differential recall).  Note that
in our case there is no a priori ``correct'' value for the similarity
bias parameter. Different values lead to different tradeoffs between
the differential precision and recall, and the right operating point
depends on the application and even on the analysis stage (similar to
ROC analysis in standard classification). In contrast, in the usual
use of transfer learning where the goal is to recover the individual
networks, there is a ``correct'' value for this parameter that depends
on how similar the true networks are (if the similarity bias is too
strong performance will drop due to negative transfer, while if the
similarity bias is too weak not enough useful information is
transferred between tasks.).

While the tradeoff between precision and recall for learned
differences is mainly controlled by the $\lambda_2$ parameter, the
sparsity parameter, $\lambda_1$, also has an effect on the differences
learned because it controls which edges are present in each
network. Figure~\ref{fig:sparsity} shows the differential
precision-recall tradeoff for various values of $\lambda_1$. Lower
values of $\lambda_1$ (denser networks) increase the differential
recall by identifying differences due to weaker dependencies that do
not appear in the sparser graphs. However, the differential precision
is lowered because more fake edges are learned in each graph which
induce fake differences. At the extreme when the networks are too
dense ($\lambda_1 = 0.1$), there are ten times more spurious edges
than real ones so the differential precision is low even when the
similarity bias is high.

We also vary the rewiring probability when generating the true
networks, varying the number of true
differences. Figure~\ref{fig:pmove} shows the differential
precision-recall graphs for various values of the rewiring probability
for $\lambda_1=0.4$. As the fraction of true differences
decreases (lower rewiring probability), it gets harder to identify
them and performance drops. Increasing the similarity bias (increasing
$\lambda_2$), however, still leads to higher differential
precision. These results highlight another fundamental difference with
the usual use of transfer learning. In transfer learning the true
networks must be similar in order to get any benefit, while in
differential analysis there is no such constraint. In fact, in
differential analysis performance improves if the true networks are
more dissimilar, as opposed to transfer learning where performance gets
better with more similar true networks.

Finally, we compare to using bootstrap procedures to identify higher
confidence differences. For the bootstrap procedure, we generate a
bootstrap sample of the data, train independent graphical models on
that data, then compare the learned networks. We repeat this 44400
times. For each edge, $e$, we calculate the bootstrap frequency,
$P_B(e)$, of it appearing in one network but not the other (i.e. $e$
is a difference).  For a given cut-off, $c$, we consider all edges
with $P_B(e) \geq c$ as inferred differences. We then generate a
differential precision-recall graph by varying $c$ from 0 to 1 so that
at $c=0$ any difference that appeared in any bootstrap is considered a
difference, and at $c=1$ only differences that appear in all 500
bootstraps are considered differences. Figure~\ref{fig:bootstrap}
shows the comparison between the transfer method and the bootstrap
method for $\lambda_1=0.4$. The transfer method dominates the
precision-recall curve compared to bootstrapping, and, importantly,
can reach a high precision regime that is unattainable via
bootstrapping. Also, the transfer method requires about a factor of 50
less computation than the bootstrapping (the transfer method learns
the networks about 10 times, once for each setting of $\lambda_2$,
while bootstrapping learns the networks 500 times, once for each
bootstrap). Moreover, the bootstrapping procedure becomes increasingly
computationally expensive as higher levels of differential precision
(or recall) are desired. Bootstrapping achieves the highest
differential precision when $c=1$ and there are $\sim$100 differences
($\sim$30 of them false) that occur in all 500 bootstraps. Therefore,
to push the precision ratio higher than 0.7, we would need to run more
bootstraps until these false differences do not appear in at least one
of them so they can be filtered out (while hoping that some true
differences remain).

\section{Real Case Studies}
 \label{sec:fdr}
In this section we present three real case studies where domain
experts performed differential dependency network analysis in an
ovarian cancer study, a pancreatic cancer study, and a neuroimaging
study. A quantitative evaluation of the results is difficult in real
usage scenarios because there is no ground truth to compare against so
true differential precision and recall can not be calculated. In these
cases, the usual approach is to estimate the false discovery rate
(FDR) through permutation tests. We take the following approach to
estimating FDR: first pool the data from all populations (conditions)
together, then randomly split the data into synthetic populations with
the same number of instances as the original ones. There should not be
any difference between the dependency structures of the newly
generated synthetic populations, so any difference identified by the
algorithm is a false discovery. The splitting procedure is performed
multiple times and the average FDR is used as an estimate of the FDR
of the algorithm on the original problem. This approach obviously is
not perfect and can underestimate the FDR. When tested on synthetic
data we found that the true FDR is indeed underestimated.
 
When learning dependency differences from data, we have observed from real users that the best use of multi-network learning is as a part of an exploratory tool that allows interactive exploration of the various tradeoffs controlled by the sparsity and transfer parameters. To this end, we provide the domain experts with a Cytoscape \cite{Shannon:2003uq} plugin that
allows them to interactively explore different settings of $\lambda_1$
and $\lambda_2$ and quickly identify and visualize the differences in
the dependency networks (see Appendix~\ref{app:cytoscape}).
For the most part there is no ``correct'' setting of the
parameters as different tradeoffs convey different information about
the domain and the right operating point changes depending on the
application and even on the analysis stage.

\subsection{Oncology Studies}
Domain experts applied the technique described in this paper to
perform differential dependency network analysis with the goal of
identifying and analyzing cancer-induced changes in the dependency
structure of plasma proteins. They analyze two cancers: ovarian and
pancreatic. The ovarian cancer study uses data from a cohort of 247
patients (114 cases and 133 controls). The pancreatic cancer study
uses a cohort of 469 patients (239 cases and 230 controls).  Each
patient had a blood sample taken prior to the diagnosis, and plasma
concentrations of 858 proteins were measured using Somamer technology
\cite{Gold:2010fk}.

\begin{figure}[tb]
\centering
\subfloat[Ovarian Cancer]{\includegraphics[width=0.48\columnwidth]{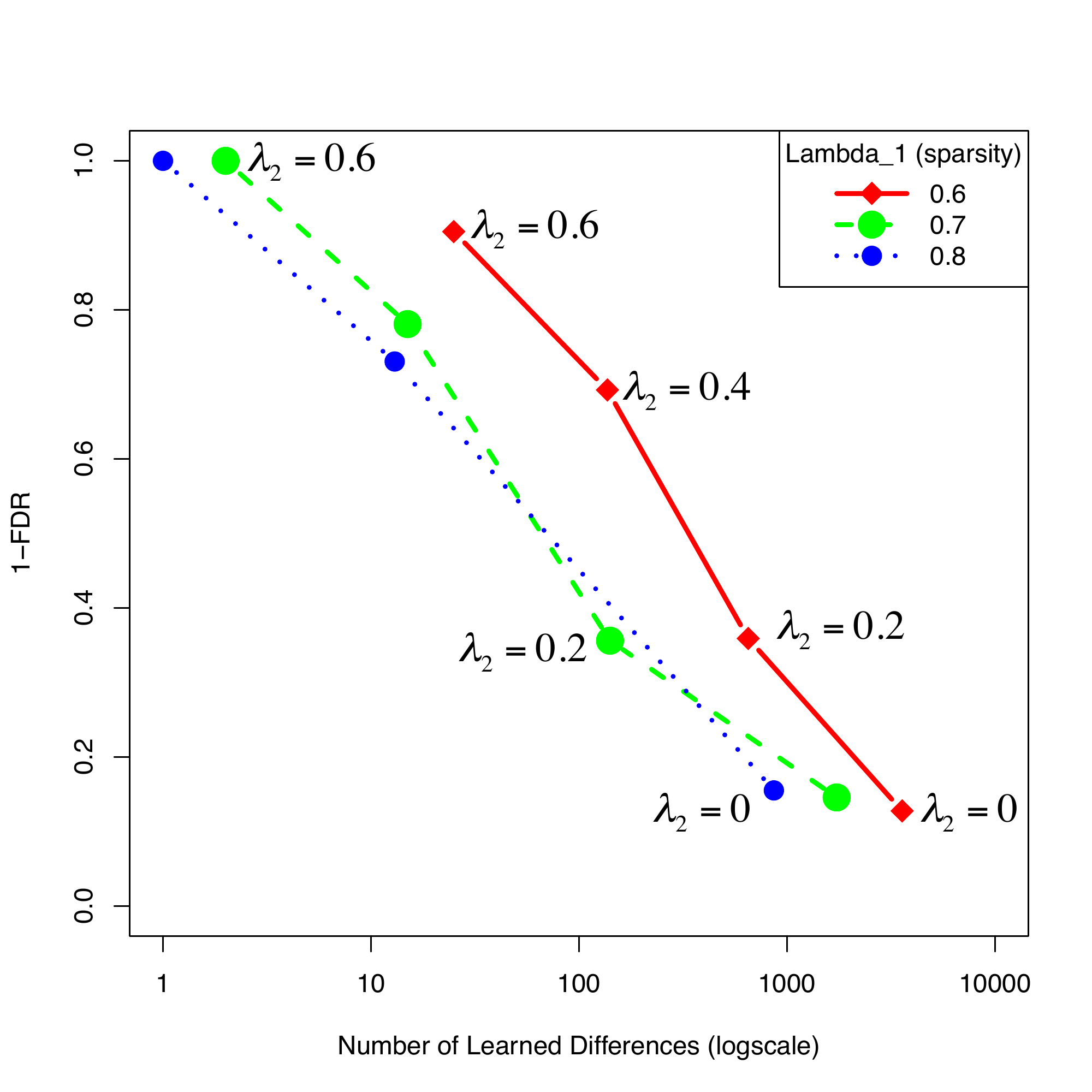}
\label{fig:ovarianNumVsRatio}} \hfil
\subfloat[Pancreatic Cancer]{\includegraphics[width=0.48\columnwidth]{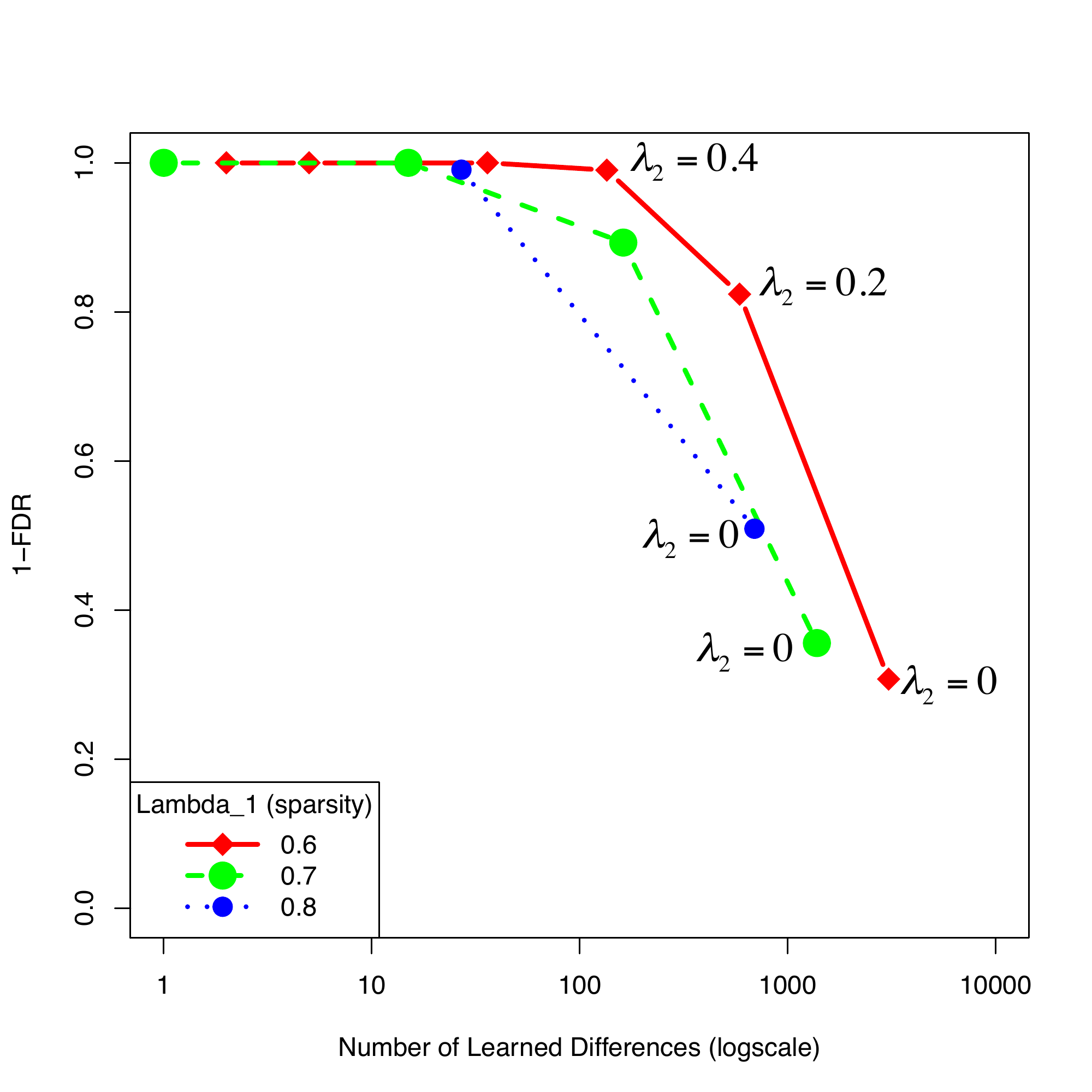}\label{fig:pancreaticNumVsRatio}}
\caption{Oncology studies. Tradeoff between FDR and number of differences.}
\label{fig:realstudiesNumvsRatio}
\end{figure}

Analogous to the precision recall curves in
Figure~\ref{fig:synthetic_precision},
Figure~\ref{fig:realstudiesNumvsRatio} shows the tradeoff between
estimated differential precision (1-FDR) and the number of differences
found (in log-scale) for the Ovarian and Pancreatic cancer studies. In
both cases, the standard approach of learning the networks
independently and comparing them ($\lambda_2 = 0$ at the right end of
the curves) discovers between 1000 and 5000 differences, but the
majority of them (almost 90\% for Ovarian and more than 50\% for
Pancreatic) are estimated to be false. This level of false discovery
renders the results of the differential analysis pretty much
useless. However, as the bias for learning similar networks ($\lambda_2$)
increases, the estimated FDR steadily decreases for all settings of the
sparsity parameter, reaching levels below 10\% which is very
acceptable in biological applications.  As in the synthetic data,
fewer differences are found, but we have much higher confidence that
the remaining differences are real.

\begin{figure}[tb]
  \centering
  \includegraphics[width=0.8\columnwidth, bb=89 350 532 751]{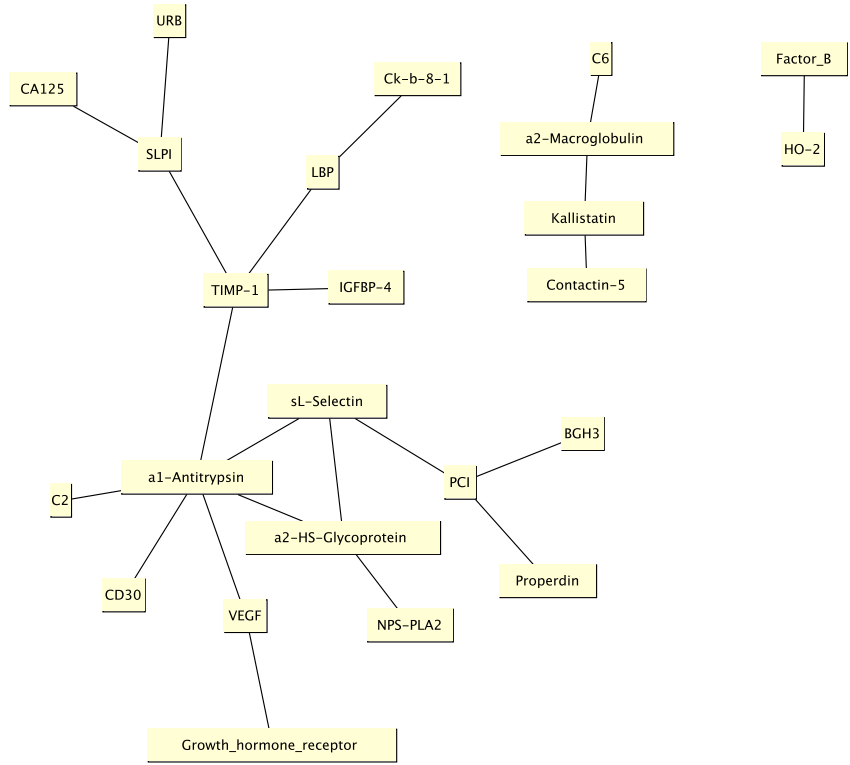}
  \caption{Differential dependency network between Case and Control populations in the Ovarian oncology study. Each edge represents a dependency that is present in the cancer population but not in the control population.}
  \label{fig:ovarianNet}
\end{figure}

Figure~\ref{fig:ovarianNet} shows the differential dependency network
between the cancer and control populations in the Ovarian study for
$\lambda_1=0.6$ and $\lambda_2=0.6$. Every arc in this network
represents a dependency that is present in the cancer population but
not in the control population. For comparison, we also show the
differential dependency network obtained using the standard technique
($\lambda_1=0.6$, $\lambda_2=0$) in Figure~\ref{fig:ovarianNetNoTran} in Appendix~\ref{app:ovarian}. To ensure that the
differential network in figure~\ref{fig:ovarianNet} reveals relevant
biological information, we ran a standard enrichment analysis using
DAVID \cite{Huang:2008fk} on the 24 proteins that appear in the
figure, and asked collaborators with extensive expertise in cancer
biology to analyze the results. The enrichment analysis shows that the
following functional clusters are significantly enriched\footnote{A
  functional cluster is enriched if there are significantly
  more members of that cluster present in the query list than it would
  be expected from the random background distribution.}: endopeptidase
inhibitor, inflammatory response, complement and coagulation, and
extracellular matrix. This is consistent with what is known about
ovarian cancer biology. The body's reaction to ovarian cancer includes
stimulation of both the adaptive (antibodies, cellular immunity) and
innate (complement, inflammation) immune systems. In fact, the new
``foreign" entity (ovarian cancer) that stimulates these responses
also creates a new milieu in which tumor mutations are selected for
when they help the cancer evade these immune responses
\cite{wang2005ovarian}. Ovarian cancers (as well as many other
cancers) also tend to induce a hyper-coagulable state, which involve
coagulation and complement proteins. Endopeptidases play essential
roles in homeostasis and signal transduction.  Changes in the
extracellular matrix are also key to the process as cancer cells
escape the primary tumor and metastasize. A list of proteins
associated with each of these processes is given in Appendix~\ref{app:ovarian}. Many of the proteins in Figure~\ref{fig:ovarianNet} have
been associated with cancer in general and with ovarian cancer in
particular. For instance CA125 is a well known and clinically used
ovarian cancer marker; SLPI has been shown to be over-expressed in
gastric, lung and ovarian cancers, accelerating metastasis
\cite{choi2011secretory}; VEGF is involved in the growth of blood
vessels\footnote{Tumors require heavy vascularization to grow.};
IGFBP4 has been associated with a number of cancers, including ovarian
\cite{walker2007insulin};

\begin{figure}[tb]
  \centering
  \includegraphics[width=\columnwidth, bb=38 372 576 742]{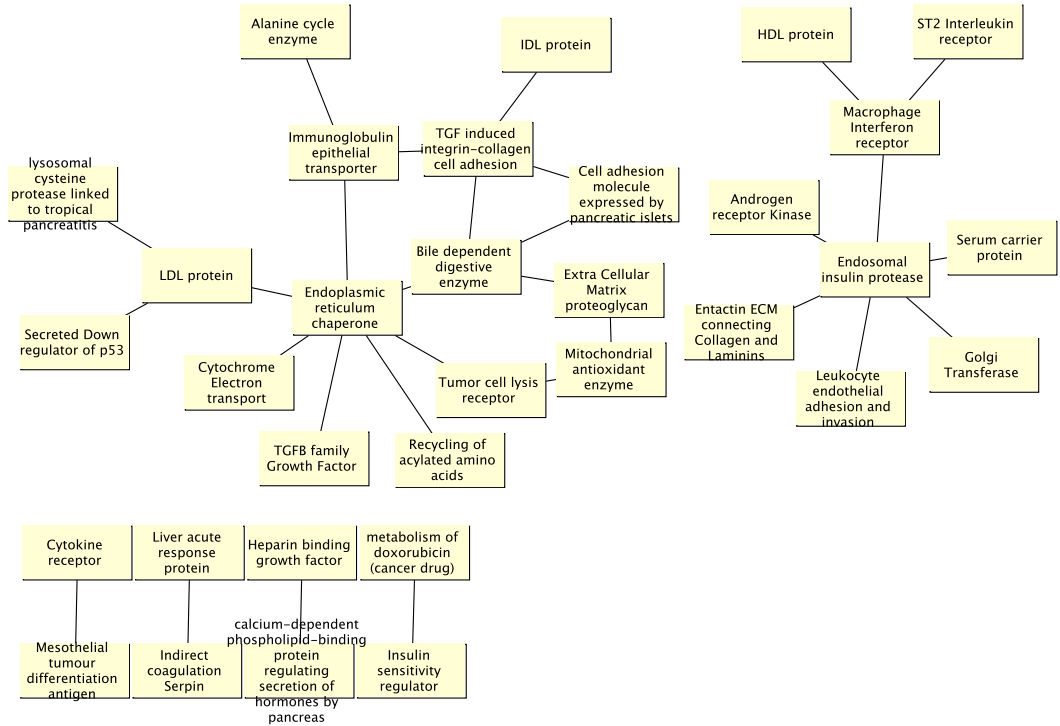}
  \caption{Differential dependency network between Case and Control populations in the Pancreatic oncology study. Each edge represents a dependency that is present in the cancer population but not in the control population.}
  \label{fig:pancreaticNet}
\end{figure}

Figure~\ref{fig:pancreaticNet} shows the differential dependency
network between the cancer and control populations in the pancreatic
cancer study for $\lambda_1=0.6$ and $\lambda_2=0.6$, with the node
labels showing functional descriptions in lieu of the protein
names.\footnote{Since the results could be of significant commercial
  interest in pancreatic cancer diagnosis, our collaborators requested
  that we do not reveal the actual proteins in this network until a
  patent is filed to protect the IP. By the time of publication this
  should not be an issue any more and the actual protein names will be
  revealed.} The differential dependency network shows proteins linked
with the endocrine pancreas (e.g. endosomal insulin protease, insulin
sensitivity regulator, protein regulating secretion of hormones by
pancreas) and with the exocrine pancreas (e.g, HDL, LDL, IDL proteins,
bile dependent digestive enzyme), as well as proteins associated with
cancer and cancer related processes (e.g. tumor cell lysis receptor,
mesothelial tumor differentiation antigen, down regulator of p53,
endoplasmic reticulum chaperone). An enrichment analysis finds the
following processes to be significantly enriched: extracellular
matrix, lipid transport and cell adhesion. These processes are
relevant to the pancreatic cancer biology. As mentioned above, changes
in the extracellular matrix are involved in cancer cells escaping the
primary tumor and metastasizing. Related to the extracellular matrix,
cell adhesion is also a key process that regulates the migration
(spreading) of cancer cells through the body and the destruction of
the histological structure in cancerous tissues
\cite{hirohashi2005cell}. The lipid transport is related to the
exocrine pancreatic function \cite{lopez1993cholesterol}.

\subsection{Accelerated Learning fMRI Study}

Functional magnetic resonance imaging (fMRI) measures the activity
level in regions of the brain while a subject is in the scanner. The
dependency network between regions of interest (ROI) in the brain, is
called a functional brain network because it indicates which regions
have activity patterns that appear to be exchanging information with
each other. A common question is whether these dependencies are
different in subjects under different conditions.

Using data from the Accelerated Learning fMRI Study, we want to see
how brain regions interact before and after a person learns a new
skill \cite{Clark2012TDCS-guided-usi}. In this study, subjects are
asked to identify concealed objects in still images taken from a
virtual reality environment. Initially, all subjects are considered
Novice (i.e. not significantly better than random at identifying
images with concealed objects). fMRI data are collected from these
subjects while performing this identification task. Then, subjects are
trained until they reach a level of Intermediate competency (midway
between chance and perfect). At this point, fMRI data are again
collected while performing the identification task. In total, we have
data from 12 subjects at the Novice stage and 4 at the Intermediate
stage. For each subject, there are 1056 samples of brain activity from
116 regions of interest (ROIs) in the brain. The ROIs are defined by
the AAL atlas \cite{mni_aal_tzourio_2002}. The goal is to identify
dependencies among the ROIs that are different between the
Intermediate and Novice stages. Looking at the networks (rather than
the activity levels of individual ROIs) shows us which ROIs are most
critical for performing a cognitive task
\cite{Clark2012TDCS-guided-usi}.

\begin{figure}[tb]
\centering
\subfloat[]{\includegraphics[width=0.48\columnwidth, bb= 4 18 475 446]{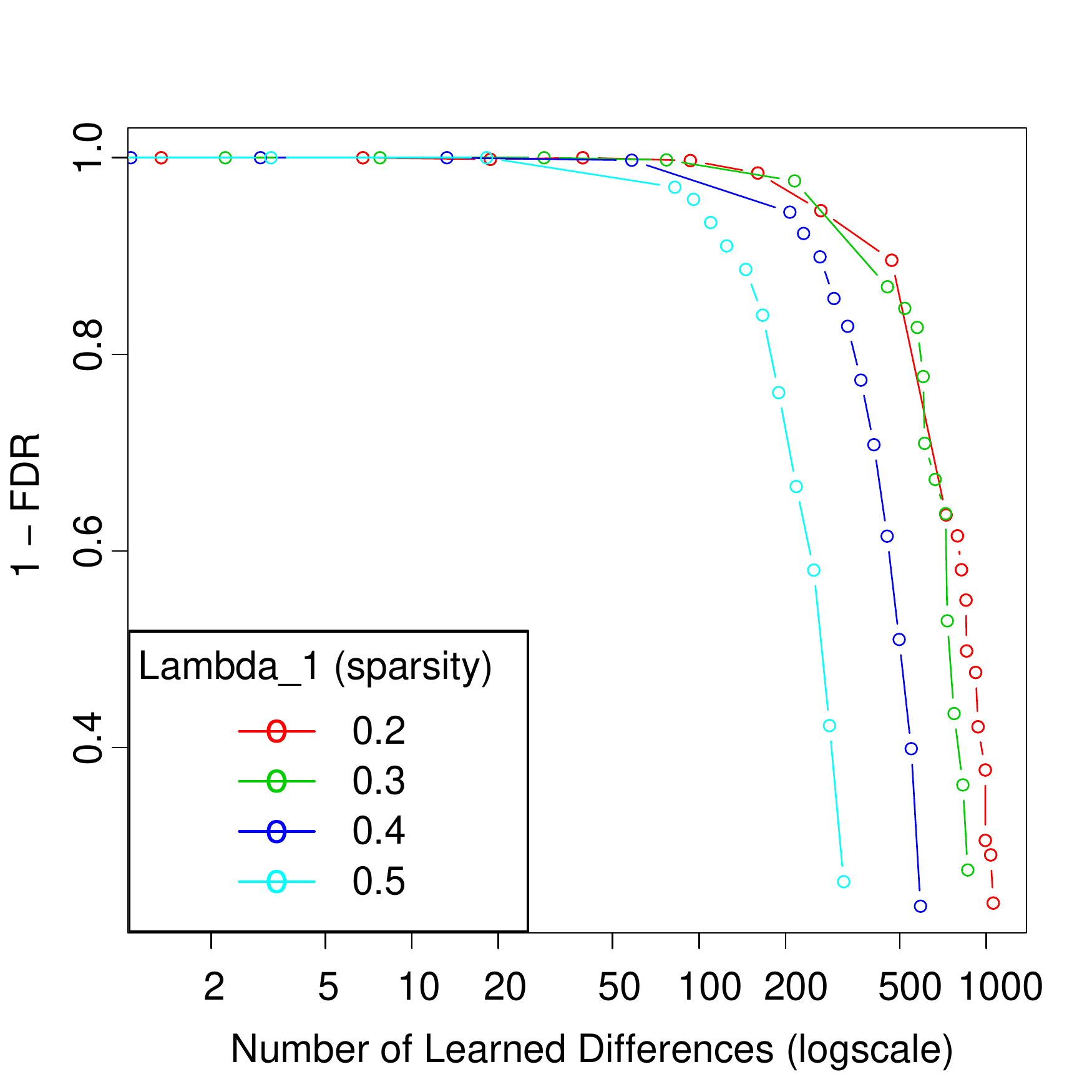}\label{fig:novintNumVsRatio}} \\
\subfloat[]{\includegraphics[width=0.48\columnwidth]{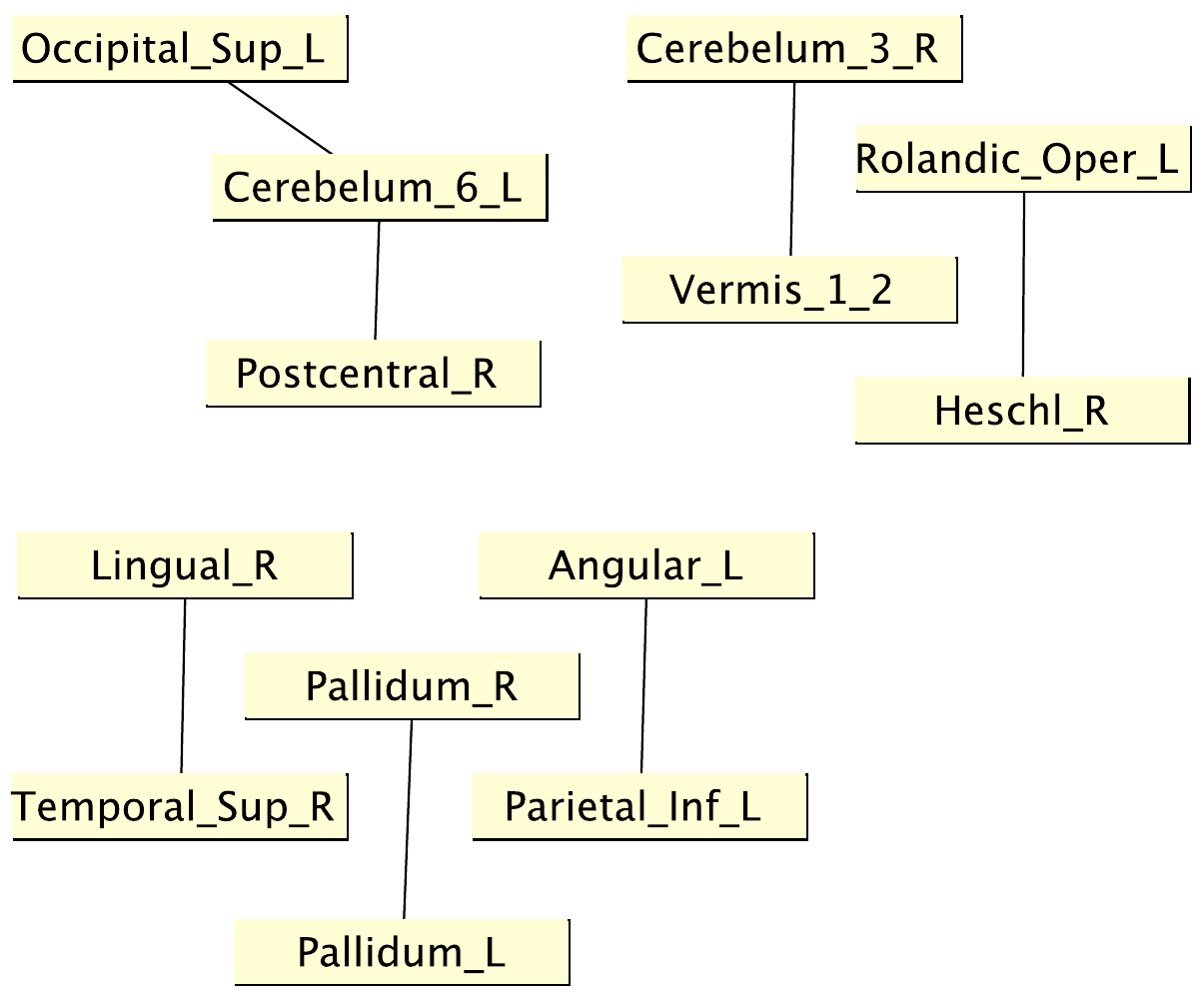}
\label{fig:darpa_net_nov}} \hfil
\subfloat[]{\includegraphics[width=0.48\columnwidth]{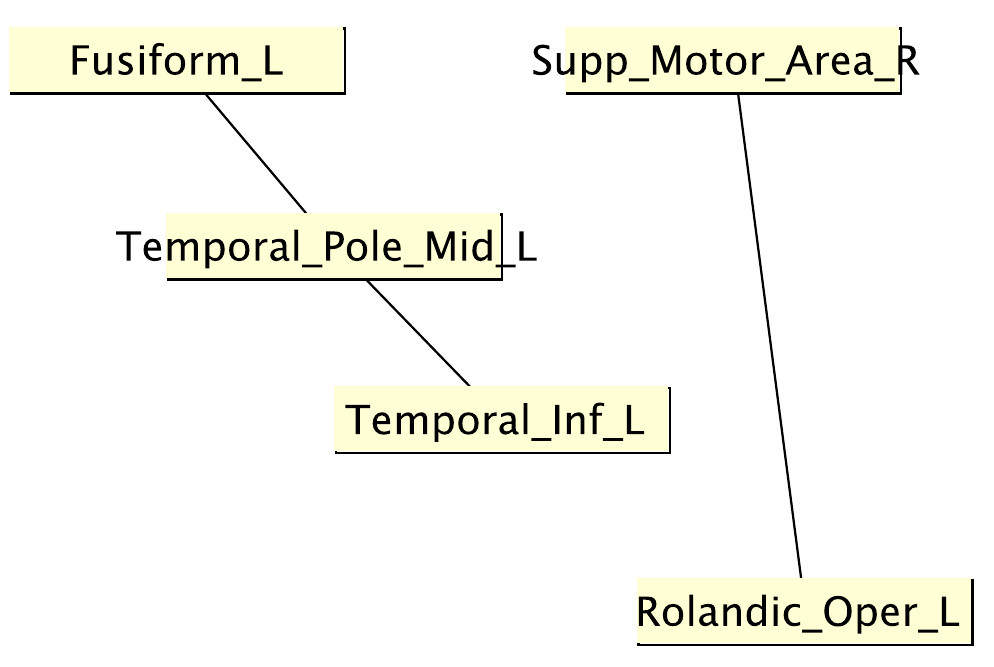}
\label{fig:darpa_net_int}} 
\caption{Accelerated Learning study. (a) Tradeoff between FDR and number of differences (b) Edges in Novice but not in Intermediate, (c) Edges in Intermediate but not in Novice}
\label{fig:darpa_nets}
\end{figure}

Figure~\ref{fig:novintNumVsRatio} shows the tradeoff between the
estimated differential precision (1-FDR) and the number of differences
found (in log-scale) for different values of $\lambda_1$. As before,
$\lambda_2$ is varied to obtain each curve. Using the standard
approach ($\lambda_2=0$, lower right end of the curves), one identifies
from 300 to 1000 differences between novice and intermediate
functional brain networks, but almost 80\% of them are estimated to be
false. As the similarity bias increases, the estimated FDR decreases
rapidly to levels close to 0 (again, remember that this is an
optimistic estimate), so we are much more confident that we are
identifying true differences in the functional brain networks.


Figures~\ref{fig:darpa_net_nov}~and~\ref{fig:darpa_net_int} show the
connections that appear in novices but not in intermediates, and vice
versa for $\lambda_1 = 0.5$ and $\lambda_2 = 0.2$ (these parameter
settings were selected by the domain expert to give just a handful of
dependencies of high confidence). Figure~\ref{fig:darpaSharedNet} in Appendix~\ref{app:fmri} also
shows the connections present in both novices and intermediates. These
results show that for both stages, groups of brain regions are found
that share information, which correlate well with sensori-motor
pathways found in humans. This includes the occipito-parietal dorsal
visual pathway that computes the location of objects, the
occipito-temporal ventral pathway that determines the identity of
objects, collections of frontal and cingulate regions that help to
make decisions about responses, as well as separate cerebellar and
middle temporal networks, along with other smaller networks of brain
regions \cite{Mishkin:1983kx}.  With learning to identify hidden
objects in this task, it was found that portions of the ventral
pathway increased in strength, suggesting that learning resulted in
greater information flow among regions that specialize in visual
object identification.

\section{Conclusion}

Differential analysis of dependency networks of multivariate data
allows domain experts to uncover and understand the differences
between related populations and the processes that are generating
these differences. Such questions arise in many domains including
biology, medicine, and neuroscience. We have shown that the
traditional approach of learning the dependency networks for each task
independently and comparing them is prone to having high false
discovery rates. We have discussed the importance of controlling the
quality of the inferred differences between dependency networks,
and explored a novel use of transfer learning techniques to provide a
natural and explicit ``knob'' that controls the precision-recall
tradeoff in differential network analysis. We have shown empirically
that this approach achieves higher precision than existing methods,
and yields better performance than significantly more expensive
bootstrapping procedures. Finally, we have presented three real case
studies where domain experts used the proposed techniques to uncover
compelling evidence of biological processes involved in cancer and
human learning.

While in this paper we have focused on differential network analysis,
the idea of using transfer learning techniques to improve differential
analysis is quite general. For instance, similar techniques could be
used in conjunction with feature selection to answer questions like
``are there different cancer biomarkers for men and women?'', or in
conjunction with clustering/unsupervised learning to detect
significant changes in cluster structures between conditions.

\section*{Acknowledgements}
We would like to acknowledge the contributions of several collaborators. The pancreatic cancer samples were collected by Randall Brand, M.D. of the University of Pittsburgh Medical Center and Michelle A. Anderson, M.D. of the University of Michigan Hospital and Health Systems. Britta Singer and Ed Brody of SomaLogic Inc. helped with the analysis of the ovarian and pancreatic cancer results. 

\bibliography{MyLibrary}
\bibliographystyle{unsrt}

\clearpage
\appendix

\section{Ovarian Cancer Results}
\label{app:ovarian}

\begin{table}[h!]
\begin{small}
\begin{tabularx}{\textwidth}{| X | X | X | X | X |}
\hline
Immune response proteins & Inflammatory response proteins & Coagulation and complement proteins & Proteins that are involved in the extracellular matrix & Endopeptidase inhibitor proteins\\
\hline
a2-Macroglobulin & a2-Macroglobulin & a2-Macroglobulin & TIMP1 & SLPI \\
C2 & Ck-b-8-1 & C2 & TIMP1 & TIMP1 \\
C6 & GHR & C6 & URB & a2-Macroglobulin \\
Ck-b-8-1 & LBP & Factor B & a1-Antitrypsin & \scriptsize{a2-HS-Glycoprotein} \\
Factor B & a1-Antitrypsin & a1-Antitrypsin & BGH3 & a1-Antitrypsin \\
Properdin & TIMP-1 & PCI & VGEF & Kallistatin \\
GHR & CD30 & TIMP-1 &  & PCI \\
LBP & VEGF & CD30 & & \\
sL-Selectin & \scriptsize{a2-HS-Glycoprotein} & VEGF & & \\
a1-Antitrypsin & & & & \\
TIMP-1 & & & & \\
VEGF & & & & \\
CA-125 may also be involved in these responses. & & & & \\
\hline
\end{tabularx}
\end{small}
\caption{Proteins from the Ovarian cancer differential network that are involved in each of the enriched functional processes.}
\end{table}





\begin{figure}[h!]
\centering
\includegraphics[width=\textwidth, bb=142 337 480 750]{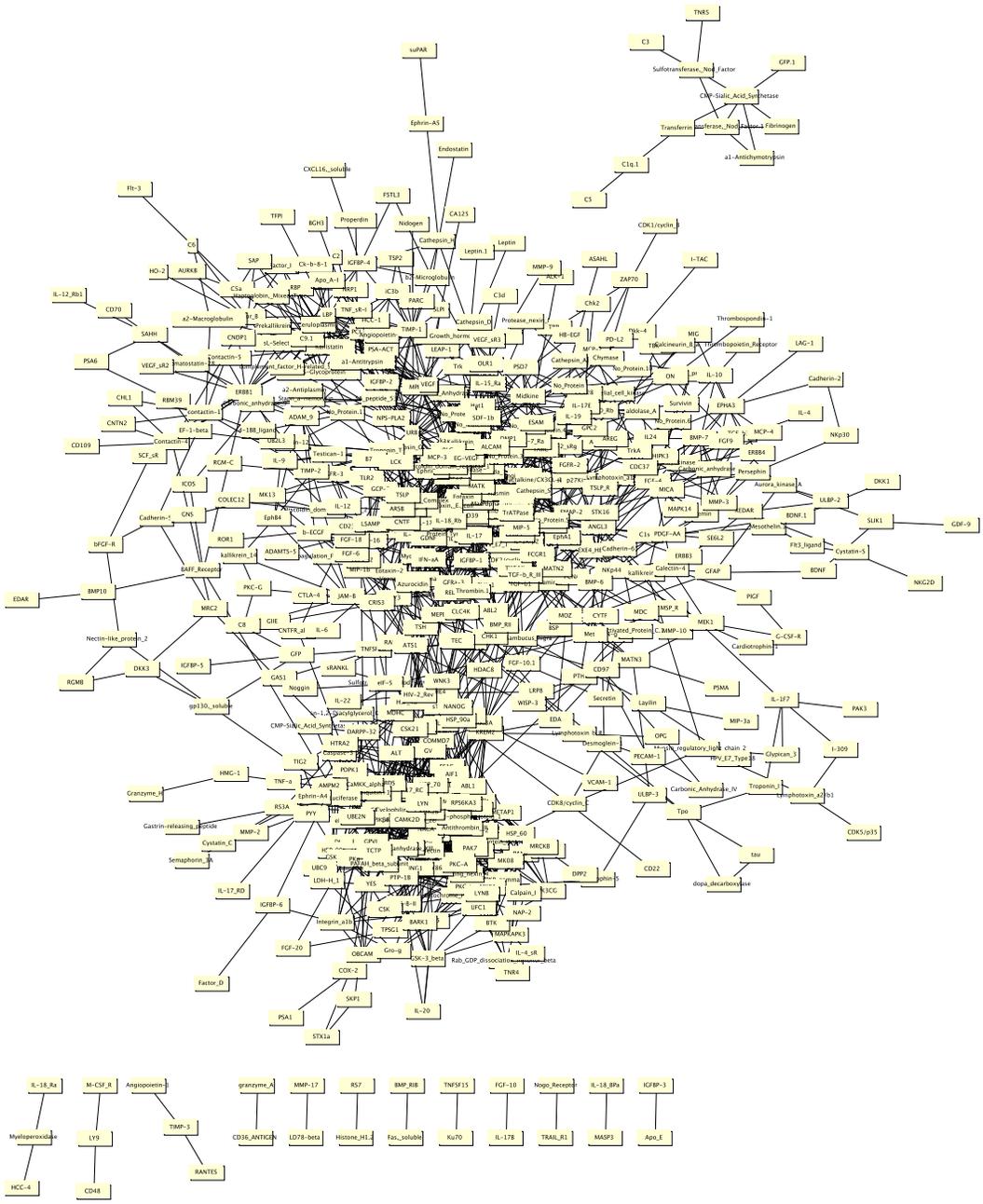}
\caption{Differential network for Ovarian cancer obtained using the traditional method of learning the dependency networks independently and compare them ($\lambda_1=0.6$, $\lambda_2=0$).}
\label{fig:ovarianNetNoTran}
\end{figure}

\clearpage
\section{Accelerated Learning fMRI Results}
\label{app:fmri}


\begin{figure}[h!]
\centering
\includegraphics[width=\textwidth, bb=131 317 457 767]{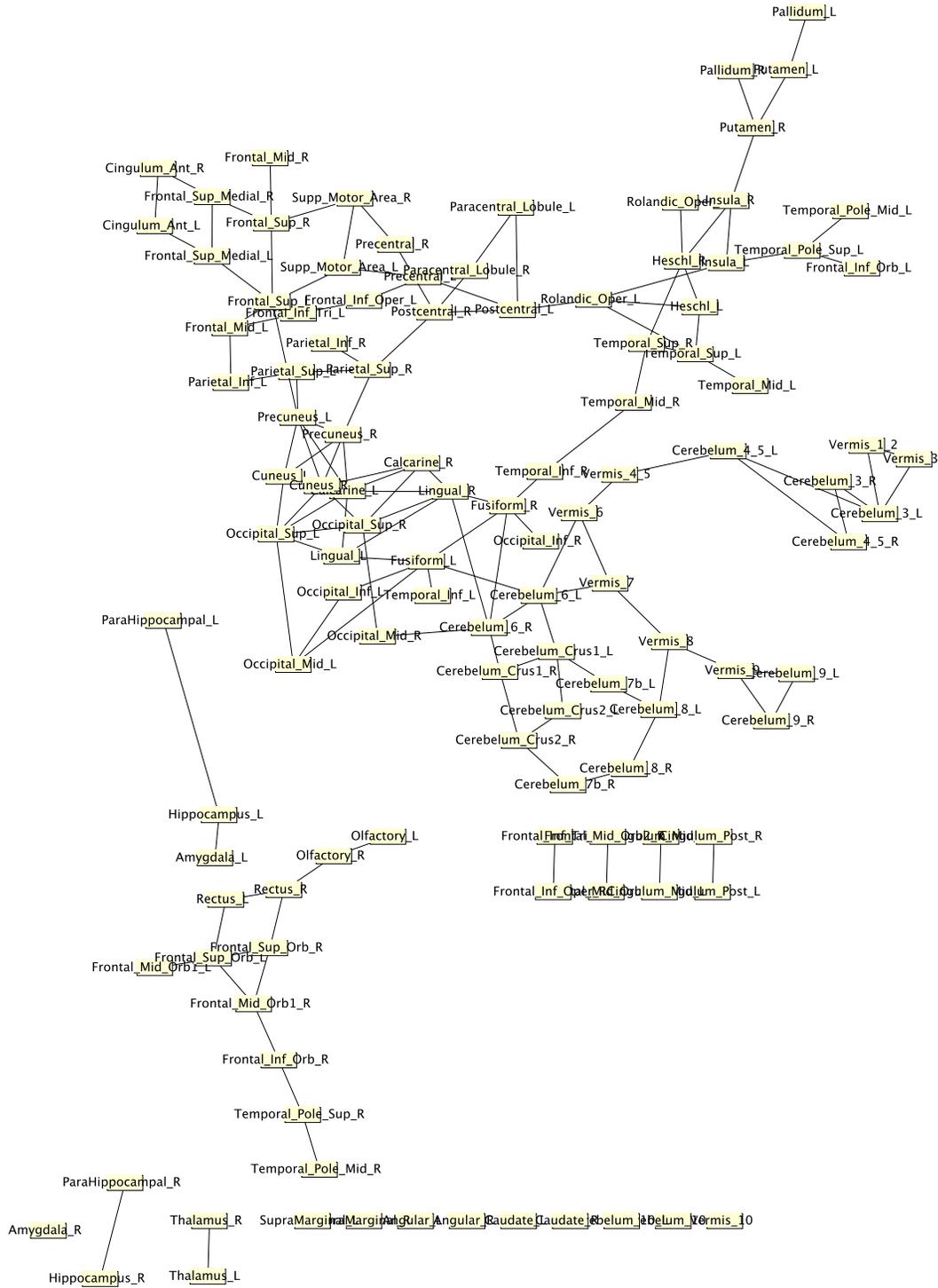}
\caption{Network of dependencies shared among Novice and Intermediate stages of the Accelerated Learning study. The network gives valuable re-assurance that the network learning algorithm is identifying true pathways.}
\label{fig:darpaSharedNet}
\end{figure}

\clearpage

\section{Interactive Visualization of Differential Dependency Networks}
\label{app:cytoscape}

\begin{figure}[h!]
\centering
\includegraphics[width=\textwidth]{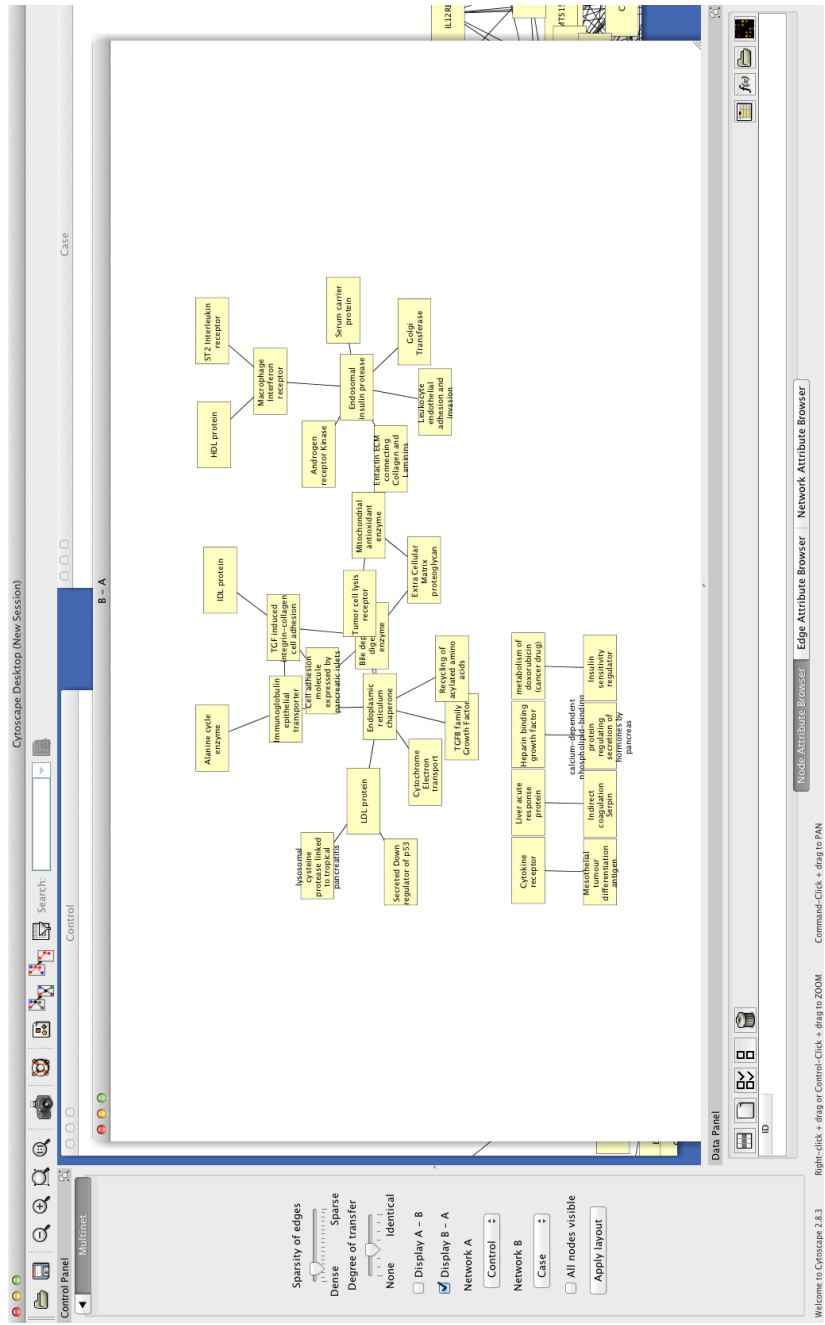}
\caption{Cytoscape plugin for interactive visualization of the differential networks and exploration of the differential precision-recall tradeoffs.}
\label{fig:plugin}
\end{figure}

\end{document}